\documentclass{article}

\PassOptionsToPackage{numbers, compress}{natbib}

\usepackage[preprint]{neurips_2021}

\usepackage{tikz}
\usetikzlibrary{shapes,arrows,bayesnet, patterns, positioning,decorations.markings}

\definecolor{blu}{RGB}{50,106,170}
\definecolor{gre}{RGB}{63,129,71}
\definecolor{drd}{RGB}{179,45,38}
\definecolor{pur}{RGB}{81,64,130}

\usepackage{wrapfig}
\usepackage[utf8]{inputenc} 
\usepackage[T1]{fontenc}    
\usepackage{hyperref}       
\usepackage{url}            
\usepackage{booktabs}       
\usepackage{amsfonts}       
\usepackage{nicefrac}       
\usepackage{microtype}      
\usepackage{xcolor}         

\hypersetup{
  colorlinks=true,
  citecolor=blu,
  linkcolor=drd,
  urlcolor=pur
}

\usepackage{float}
\usepackage{natbib}
\usepackage[ruled]{algorithm2e}

\newcommand{\figureposition}[2]{{\color{#1} \footnotesize \scshape #2}}

\title{Linear-Time Probabilistic Solutions\\ of Boundary Value Problems}

\author{Nicholas Krämer\\
  University of Tübingen\\
  Tübingen, Germany \\
  \texttt{nicholas.kraemer@uni-tuebingen.de} \\
  \And
  Philipp Hennig \\
  University of Tübingen and\\ MPI for Intelligent Systems\\
  Tübingen, Germany \\
  \texttt{philipp.hennig@uni-tuebingen.de}}

\usepackage{amsmath, amssymb}
\usepackage{bm}

\newcommand{\Ebb}{\mathbb{E}}
\newcommand{\Rbb}{\mathbb{R}}

\newcommand{\Tbb}{\mathbb{T}}

\newcommand{\Ncal}{\mathcal{N}}

\newcommand{\Ocal}{\mathcal{O}}

\newcommand{\Vcal}{\mathcal{V}}

\newcommand{\diff}{\,\text{d}}

\DeclareMathOperator{\argmax}{arg\,max}
\DeclareMathOperator{\argmin}{arg\,min}

\usepackage{amsthm}
\newtheorem{theorem}{Theorem}
\newtheorem{proposition}[theorem]{Proposition}

\newtheorem{remark}[theorem]{Remark}

\theoremstyle{definition}

\usepackage{amsmath}
\usepackage{pdfpages}

\usepackage{xcolor}
\usepackage{cleveref}
\usepackage{blindtext}

\begin{document}

\maketitle

\begin{abstract}
  We propose a fast algorithm for the probabilistic solution of boundary value problems (BVPs), which are ordinary differential equations subject to boundary conditions.
  In contrast to previous work, we introduce a Gauss--Markov prior and tailor it specifically to BVPs, which allows computing a posterior distribution over the solution in linear time, at a quality and cost comparable to that of well-established, non-probabilistic methods.
  Our model further delivers uncertainty quantification, mesh refinement, and hyperparameter adaptation. We demonstrate how these practical considerations positively impact the efficiency of the scheme. Altogether, this results in a practically usable probabilistic BVP solver that is (in contrast to non-probabilistic algorithms) natively compatible with other parts of the statistical modelling tool-chain.
\end{abstract}

\section{Boundary value problems in computational pipelines}

This work develops a class of algorithms for solving  \emph{ODE boundary value problems}; that is, ordinary differential equations (ODEs)
\begin{align}\label{eq:ode_bvp}
  \dot y(t) = f(y(t), t)
\end{align}
subject to  \emph{left-} and \emph{right-hand side boundary conditions}  $L y(t_0) = y_0$ and $R y(t_\text{max}) = y_\text{max}$.
The vector field $f: \Rbb^d \rightarrow \Rbb^d$, as well as $L \in \Rbb^{d_L \times d}, R \in \Rbb^{d_R\times d}, t_0 \in \Rbb, t_\text{max} \in \Rbb, y_0 \in \Rbb^{d_L},$ and $y_\text{max} \in \Rbb^{d_R}$ are given.
It is no loss of generality to consider a first-order boundary value problem because higher-order problems can be transformed into first-order problems \citep{ascher1995numerical}.

\begin{figure}
  \centering
  \includegraphics{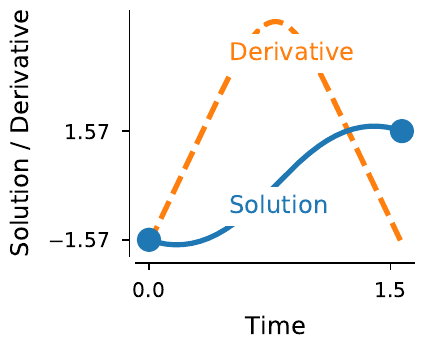}
  \includegraphics{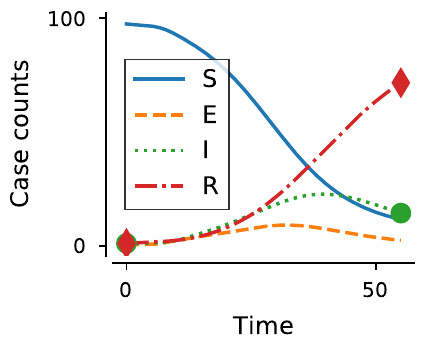}
  \includegraphics[width=0.3\textwidth]{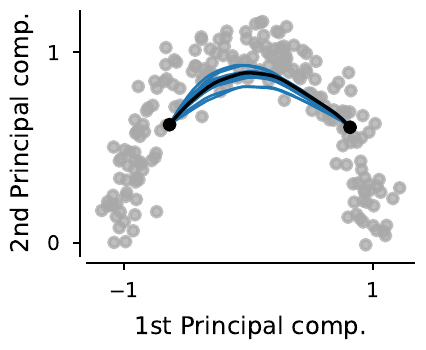}
  \caption{Recovering the trajectory of a pendulum between two positions is a BVP ({\scshape\small  left}). Lack of initial values can be made up by boundary values in an SEIR model ({\scshape\small  middle}). Straight lines on manifolds give distance measures and demand solving a BVP ({\scshape \small right}; depicted are the mean and ten samples of the probabilistic solution; principal components of 1000 MNIST images of the digit ``1'').
  }
  \label{fig:bvp_example}
\end{figure}

Loosely speaking, solving BVPs amounts to following the law of a dynamical system when ``connecting two points''. This setting is relevant to several scientific applications of machine learning.
As motivation, we consider three examples, all of which are depicted in \Cref{fig:bvp_example}.
First, recovering the trajectory of a pendulum between two positions amounts to solving the ODE $\ddot y(t) = -9.81 \sin( y(t))$ subject to the positions as boundary conditions.
If the positions were interpolated without the ODE knowledge, the output would be physically meaningless.
Second, BVPs arise when inferring the evolution of the case counts of people that fall victim to an infectious disease. A lack of counts of (a specific subset of) non-infected people at the initial time-point can be made up for by available counts of infected people at the final time-point of the integration domain.
Third, efficient manifold learning necessitates repeated computation of (geodesic) distances between two points, which amounts to solving BVPs \citep{carmo1992riemannian,arvanitidis2017latent}.
Depending on application details, the ability to produce structured output uncertainty or to enhance the algorithm by including additional sources of information can be crucial.
Probabilistic numerical algorithms respond to these challenges by solving problems of numerical simulation with probabilistic inference.
For \emph{initial} value problems, probabilistic solvers share linear-time complexity, adaptive step-size selection, and high polynomial convergence rates with their non-probabilistic counterparts \citep{kersting19,tronarp20,bosch2020calibrated,kramer2020stable}, and further provide functionality to quantify uncertainty within probabilistic programs \citep{kersting2020differentiable,schmidt2021probabilistic}.

Probabilistic BVP solvers have not yet reached this level of quality. Existing probabilistic treatments of BVPs \citep{hennig2014probabilistic, arvanitidis2019fast, john19} iteratively condition a Gaussian process on approximately ``solving the BVP''. Each such iteration requires solving a generic least-squares problem of size equal to the number of employed grid points. The resulting cubic complexity puts severe upper limits on grid resolution.
Traditional, non-probabilistic BVP solvers (for instance, those presented in \citep{ascher1995numerical}) are very efficient but do not provide probabilistic output. Thereby, they would have to serve as black-boxes inside probabilistic programs.
In this work, we close this gap. The main idea of this paper is that computing a probabilistic solution of BVPs is fast if the prior is Markovian (\Cref{sec:solve_bvp_with_probabilistic_inference}). Probabilistic modelling provides additional advantages. In particular, algorithmic parameters can be estimated automatically (including those that must be provided by the user in traditional methods; \Cref{sec:initialization_ieks}--\ref{sec:calibration_em}).

\section{Boundary value problems as probabilistic inference tasks}
\label{sec:solve_bvp_with_probabilistic_inference}

\subsection{Generative model}
\label{sec:generative_model}
Let $\sigma > 0$. We define the integrated Wiener process ${Y} = \left[Y_0, ..., Y_\nu \right]^\top: [t_0, \infty) \rightarrow \Rbb^{d(\nu + 1)}$ as the solution of the stochastic differential equation
\begin{align}\label{eq:gauss_markov_sde}
  \diff Y(t) = A Y(t) \diff t + B \diff W(t), \quad
  Y(t_0) \sim \Ncal(m_0, \sigma^2 {C}_0),
\end{align}
driven by a $d$-dimensional Wiener process $W: [t_0, \infty) \rightarrow \Rbb^d$ with diffusion $\Gamma = \sigma^2 I \in \Rbb^{d \times d}$, and initial parameters $m_0 \in \Rbb^{d(\nu + 1)}$, ${C}_0 \in  \Rbb^{d(\nu + 1) \times d(\nu + 1)}$  \citep{kersting19}.
For the moment, we set $\sigma = 1$, $m=(0, ..., 0)$, and ${C}_0 = I$, and will discuss parameters calibration later.
The entries in $A\in  \Rbb^{d(\nu + 1) \times d(\nu + 1)}$ and $B\in  \Rbb^{d(\nu + 1) \times d}$ are specified by the integrated Wiener process and imply that $Y_q$ models the $q$th derivative of the BVP solution $y$: $Y_q(t) \approx y^{(q)}(t)$, $q=0, ..., \nu$ \citep[Equation 2]{bosch2020calibrated}. This is the prior for the probabilistic BVP solver. Other choices are possible, too \citep[Section 2.1]{tronarp20}.

For ODE solvers, the likelihood is best described in terms of a measurement model \citep{tronarp19}. For BVPs, there are two sources of information: first, the \emph{boundary conditions}
\begin{align}\label{eq:likelihood_boundary_values}
  \ell_L({Y}):= L Y_0(t_0) - y_0
  \quad \text{and}\quad
  \ell_R({Y}):=R Y_0(t_\text{max}) - y_\text{max},
\end{align}
and second, the \emph{differential equation}, encoded by the information operator
\begin{align}\label{eq:likelihood_ode}
  \ell(Y)(t) := Y_1(t) - f(Y_0(t), t).
\end{align}
Similar likelihoods are used in the gradient matching literature \citep{calderhead2009accelerating, wenk2020odin}.

Let $\Tbb := (t_0, ..., t_N=t_\text{max})$ be a grid on $[t_0, t_\text{max}]$.
For now, we assume this grid is given; \Cref{sec:mesh_refinement} introduces a strategy for iterative mesh-refinement based on error-control.
We will abbreviate $\ell_n(Y) := \ell(Y)(t_n)$ and $\ell_{0:n} = (\ell_0, ..., \ell_n)$, $n=0, ..., N$.
Using $\Tbb$, as well as the likelihoods in Equations \eqref{eq:likelihood_boundary_values} and \eqref{eq:likelihood_ode}, the approximate ODE solution is captured by the posterior distribution
\begin{align}\label{eq:full_posterior}
  p\left(Y(t) \,|\, \ell_L(Y) = 0, ~\ell_{0:N}(Y) = 0, ~\ell_R(Y) = 0 \right ).
\end{align}
Unfortunately, the full posterior \eqref{eq:full_posterior} is intractable because of the non-linearity of $f$ (which implies non-linearity in all $\ell_n$).
We will thus approximate it with a Gaussian: the \emph{probabilistic BVP solution}.

\subsection{Approximate Gaussian posterior inference}

While the full posterior in \Cref{eq:full_posterior} cannot be computed in closed form, a maximum-a-posteriori (MAP) estimate is obtained by finding the minimum
\begin{align}\label{eq:map_problem_optimisation}
  \argmin_{Y(\Tbb)}\left\{-\log p(Y(\Tbb)): ~ \ell_L(Y) =0, ~ \ell_R(Y) =0, ~ \ell_{0:N}(Y) = 0\right\}.
\end{align}
This constrained optimisation problem can be solved with the iterated extended Kalman smoother (IEKS). The IEKS is a state-space implementation of a Gauss--Newton algorithm  \citep{bell1994iterated}.
As such, one step of the IEKS computes the closed-form minimum of Equation \eqref{eq:map_problem_optimisation} with a Kalman smoother, where the non-linear $\ell_{0:N}$ is replaced by its first-order Taylor approximation around the previous iterate.
Under mild assumptions on the non-linearity of $f$ and the magnitude of the objective at the optimum, Gauss--Newton methods are locally convergent with linear rate \citep{knoth1989globalization}.

Each iteration of the IEKS returns a mean and covariance function.
Eventually, the scheme approaches a variant of the Laplace approximation of the posterior (note the shorthand of \Cref{eq:full_posterior})
\begin{align} \label{eq:map_estimate_with_shorthand}
  Y_\text{\tiny MAP}(t)\sim \Ncal( m_\text{\tiny MAP}(t),  C_\text{\tiny MAP}(t)) \approx p(Y(t) \,|\, \ell_L, \ell_{0:N}, \ell_R),
\end{align}
(this is a non-standard Laplace approximation in so far as it employs a Gauss--Newton approximation of the Hessian).
A more detailed explanation is in \Cref{app:ieks_as_laplace}.
The mean $m_\text{\tiny MAP}(t)$ is the MAP estimate, because it minimises the objective in \Cref{eq:map_problem_optimisation}.
The covariance $C_\text{\tiny MAP}(t)$ is the inverse (approximate) Hessian of the negative log-posterior distribution, evaluated at $m_\text{\tiny MAP}(t)$.

The Gaussian posterior returned by the IEKS is a probabilistic BVP solution.
Thus, this basic version of the algorithm is already a valid BVP solver.
But some degrees of freedom remain, whose efficient selection improves performance significantly.
These will be the concern of the remainder of this work. Table \ref{tbl:degrees_of_freedom} presents an outline.
\begin{table}[h]
  \centering
  \caption{Configuration of the remaining degrees of freedom.}
  \label{tbl:degrees_of_freedom}
  \begin{tabular}{lll}
    \toprule
    \bf What?                       & \bf How?                            & \bf  Where?                           \\
    \midrule
    Initialisation of the IEKS      & ODE filter with Gaussian bridge     & Section \ref{sec:initialization_ieks} \\
    Mesh $\Tbb$                     & Error control                       & Section \ref{sec:mesh_refinement}     \\
    Diffusion $\sigma$              & Quasi-maximum likelihood estimation & Section  \ref{sec:calibration_em}     \\
    Initial parameters $m_0$, $C_0$ & Expectation-maximisation            & Section \ref{sec:calibration_em}      \\
    \bottomrule
  \end{tabular}
\end{table}

\section{An initial guess is not strictly necessary}
\label{sec:initialization_ieks}
Like every optimisation algorithm, the IEKS needs appropriate initialisation.
Not only does the number of iterations depend on the proximity of the initial guess to the optimum, but BVPs often allow multiple solutions, and the algorithm can find only one of them  \citep[p. 10]{kierzenka2001bvp}.
Non-probabilistic solvers outsource this issue to the user by expecting that an initial guess is provided.\footnote{For example, at the time of this writing, the BVP solvers in SciPy, Matlab, and {DifferentialEquation.jl} require the user to pass a vector of initial guesses of the solution at an initial grid to the algorithm \citep{scipybvpsolverdocs,matlabbvpsolverdocs,juliabvpsolverdocs}.}
While the same strategy is available for the probabilistic solver, there are natural alternatives in non-iterative Gaussian smoothers (Section \ref{sec:initialise_with_ode_filter}), which further benefit from combination with a bridge prior (Section \ref{sec:gaussian_bridge}).

\subsection{Initialisation with an extended Kalman smoother}
\label{sec:initialise_with_ode_filter}

If the target of a Laplace approximation of the BVP posterior is relaxed to only \emph{some} Gaussian approximation, an initial guess can be computed with an extended Kalman smoother (EKS) \citep{tronarp19, tronarp20}.
Like the IEKS, the EKS linearises the non-linear ODE measurements $\ell_{0:N}$ with a first-order Taylor series.
It differs from the IEKS in the position around which the approximation is constructed.
The IEKS linearises all $\ell_{0:N}$ at once after each completed forwards-backwards pass. The EKS linearises each $\ell_n$ on the fly during the forward pass, at the respective predictive mean \citep{sarkka2013bayesian}.
In other words, the EKS does not need an initial guess, which is why it is the tool of choice to construct one \citep{tronarp20}.

If the BVP is linear, the EKS computes the true posterior  \citep{sarkka2013bayesian,sarkka2019applied}.
If the BVP is non-linear, the EKS introduces a significant linearisation error wherever the predictive distribution deviates strongly from the true posterior.
Unfortunately, in its standard implementation, the EKS necessarily starts with incomplete information about the state $y(t_0)$ and higher-order derivatives (initialisation of which is crucial to probabilistic initial value problem solvers as well \citep{kramer2020stable}).
Ensuring that the prior distribution satisfies the boundary conditions by construction solves this problem because the iteration can never drift too far away from the optimum. The following Section \ref{sec:gaussian_bridge} explains more.

\subsection{Changing the order of updates to build a bridge prior}
\label{sec:gaussian_bridge}

\begin{figure} \centering
  \includegraphics[width=0.48\textwidth]{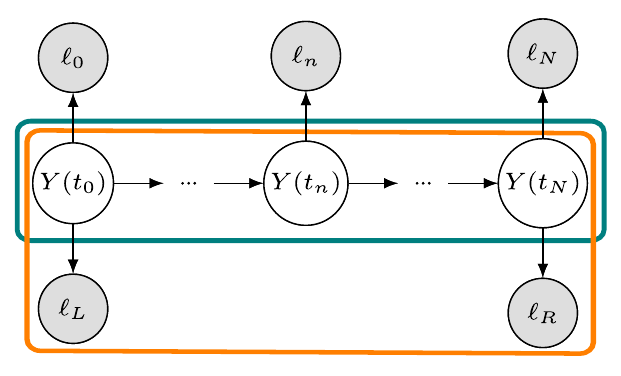}
  \includegraphics[width=0.48\textwidth]{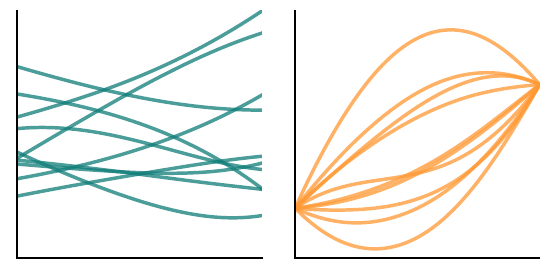}
  \caption{\textit{Construct a bridge by considering boundary values first.}
    Graphical depiction of the inference problem (\figureposition{black}{left}). Samples from the Gauss--Markov (\figureposition{teal}{centre}) and bridge prior (\figureposition{orange}{right}).
  }
  \label{fig:bridge_prior_setup}
\end{figure}

Recall that there are three sources of information: the left-hand side boundary condition $\ell_L$, the right-hand side boundary condition $\ell_R$, and the ODE measurements $\ell_{0:N}$.
If the initial and terminal state of the prior distribution are forced to accommodate $\ell_L$ and $\ell_R$ before conditioning on $\ell_{0:N}$, samples from the resulting Gaussian bridge satisfy the boundary conditions by construction; see Figure \ref{fig:bridge_prior_setup}.
The linear-time complexity of Gaussian filtering/smoothing is preserved through this change in the order of updates, because the Markov property of $Y$ yields
\begin{align} \label{eq:prior_bridge_distribution}
  p(Y(\Tbb) \,|\, \ell_L, \ell_R)=\ p(Y(t_0) \,|\, \ell_L, \ell_R) \prod_{n=0}^{N-1} p(Y(t_{n+1}) \,|\, Y(t_n), \ell_R).
\end{align}
The transition densities $ p(Y(t_{n+1}) \,|\, Y(t_n), \ell_R)$
as well as the initialisation $ p(Y(t_{0}) \,|\, \ell_L, \ell_R)$
are available in closed form (Appendix \ref{app:transition_densities_bridge}).
A reader familiar with the prediction-correction nature of Gaussian filtering can think of the implementation as follows:
Roughly speaking, each prediction step of the EKS with a bridge prior involves extrapolating from the current state to the terminal state, conditioning on the boundary condition $\ell_R$, and smoothing back to the current state.
Therefore, the computational complexity of an EKS forward-backward pass with the bridge prior is about twice as large compared to an EKS forward-backward pass with the conventional prior.
Precise derivations are in Appendix \ref{app:transition_densities_bridge}.

Figure \ref{fig:bridge_filter_initialization} shows that this extra cost is made up for by the improved linearisation behaviour because encoding the boundary conditions into the prior improves the initialisation drastically.
Following the forwards-backwards pass with the EKS, the IEKS requires only a few more iterations to find a fixed point similar to the truth.
Not using either the bridge prior or the EKS results in an initial guess that takes more iterations to find a fixed point of a lower approximation quality.
\begin{figure}
  \centering
  \includegraphics{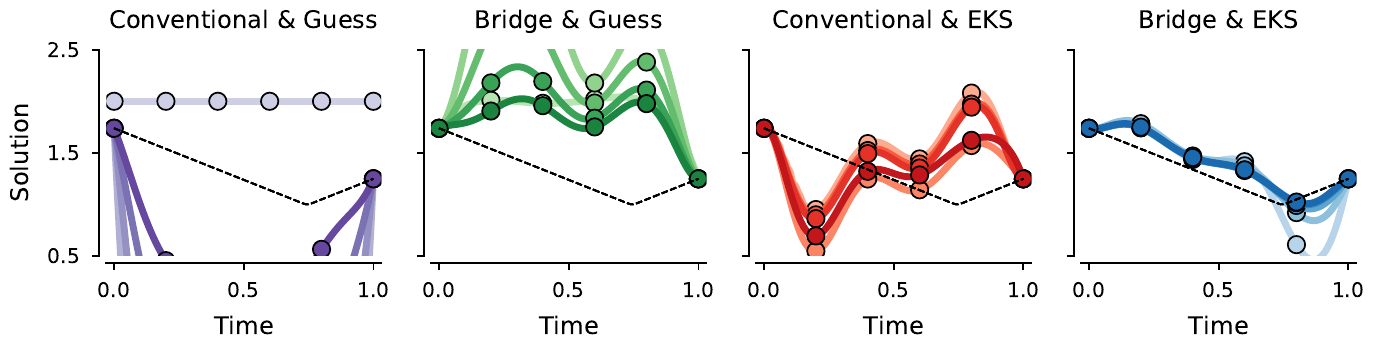}
  \caption{\textit{In combination, EKS and bridge prior initialise well.}
    Initialisation and five iterations of the IEKS depicted from light to dark on the 20th problem in \citep{mazzia2014test} (truth in black).
    Without bridges, and with an initial guess of constant twos, the fixed-point of the Gauss--Newton scheme is inaccurate on $N=6$ points (\figureposition{violet!85!black}{left}). Using either a bridge prior (\figureposition{green!50!black}{centre left}) or the EKS (\figureposition{red!75!black}{centre right}) lessens this problem.
    The bridge/EKS combination finds an accurate estimate almost immediately (\figureposition{blue!50!gray}{right}), because due to the bridge prior, the EKS linearises around a more accurate location than it would with a conventional prior during the first forward-pass.
  }
  \label{fig:bridge_filter_initialization}
\end{figure}
Abandoning both options, which aligns with initialisation of traditional BVP solvers, is least efficient since it converges to an inaccurate fixed-point.

Linear BVPs undo the effect of the bridge prior because the full posterior is computed accurately with a conventional Kalman smoother \citep{sarkka2013bayesian}.
Likewise, an IEKS iteration linearises all $\ell_{0:N}$ at once, outside of the forwards-backwards pass, which renders the bridge obsolete as well.
In other words, the changed order of updates is only relevant for the initialisation.

Of course, a fixed point of the IEKS is not necessarily a reliable BVP solution: its accuracy depends on the number and distribution of mesh points.
The following \Cref{sec:mesh_refinement} develops a principled and probabilistic approach to error control and mesh refinement in the BVP solver.

\section{Estimate the error and refine the mesh}
\label{sec:mesh_refinement}
So far, the mesh $\Tbb$ was assumed as given.
The larger the size of this mesh is, the more accurate the solution becomes; but computational cost grows linearly with the mesh size. Low error tolerances thus require smart meshing via error control. There are two (plus one) natural candidates for error estimators, all of which connect to the probabilistic formulation of solving BVPs.

\textbf{Standard deviation:}
The output of the IEKS is a Gaussian process, which can be evaluated at any point in the domain of the boundary value problem \citep[Chapter 10]{sarkka2019applied}. Its associated standard deviation provides an error estimator.
The advantage over the alternatives explained below is that it comes (essentially) for free as part of the dense output of the posterior. A potential downside of this intrinsic error estimator is its dependence on the calibration of a hyperparameter (more on this in Section \ref{sec:calibration_em}).

\textbf{Residual:}
The inference problem (\Cref{eq:full_posterior} and \Cref{fig:bridge_prior_setup}) is constructed by conditioning the prior $Y$ on attaining consistently small values in its residual $\ell(Y)(t) = Y_1(t) - f(Y_0(t),t)$. Recall that if $Y_0$ were the true ODE solution, and $Y_1$ were its derivative, $\ell(Y)$ would be zero on the whole domain.
Thus, the residual of the posterior mean of the approximate ODE solution estimates the error, which is a common approach in traditional, non-probabilistic algorithms as well (for instance \citep{kierzenka2001bvp} or \citep[Section 9.5.1]{ascher1995numerical}).
On a side note, considering the full posterior \emph{distribution} implies that the residual would be a deterministic transformation of a random variable. Thus -- in principle -- a random variable might make a more appropriate model for the residual error than a point estimate (see Remark \ref{rem:probabilistic_residual}). However, this quantity will reveal itself as inaccurate in the benchmarks below.

\begin{remark}\label{rem:probabilistic_residual}
  For a Gaussian process posterior $Y$, the law of $\ell(Y)$ is intractable in general. Linearisation of $\ell$ (at the previous iterate, like in the IEKS) unlocks a Gaussian approximation: denote the Gaussian random variable $Z(t) \approx \ell(Y)(t)$.
  An upper bound of the probability of $\|Z\|$ exceeding some tolerance,
  \begin{align}
    p\left(\|Z(t)\|^2>\text{tol}^2\right)<\left( \text{Trace}[\text{Cov}(Z(t))] + \|\Ebb(Z)(t) \|_2^2\right)/ \text{tol}^2,
  \end{align}
  is due to the Markov inequality and a third approach to error control.
  The numerator of the right-hand side will be treated as an error estimator in the benchmarks below.
  The main difference to the point estimate is that the probabilistic version punishes magnitude  \emph{and uncertainty} in the residual.
\end{remark}

All three options (which we denote by a generic $e$ from now on) estimate the error at a given $t$.
For mesh refinement, however, it is more instructive to consider the accumulated error on each interval
\begin{align}
  \epsilon_n := \left(\int_{t_n}^{t_{n+1}} \|e(t)\|_2^2 \diff t\right)^{1/2}, \quad n=0, ..., N-1.
\end{align}
If each $\epsilon_n$ is sufficiently small, the BVP solution is adequately accurate and the mesh appropriately fine.
On those intervals where $\epsilon_n$ is too large, we introduce new grid points as follows.
Assuming that the integrated error is of order $\rho>0$, $\epsilon_n \in \Ocal(h^\rho)$, splitting the interval into two equally large parts reduces the error by a factor $2^{-\rho}$, and splitting it into three equal parts by a factor $3^{-\rho}$. We use these threshold values to guide where to introduce one point and where to introduce two points.
Like \citet{kierzenka2008bvp}, we never introduce more than two at once.
For the experiments herein, and $\nu$-times integrated Wiener processes, we use $\rho = \nu + 1/2$ (which has not been proved yet but seems like a reasonable conjecture in light of Theorem 3 of \citet{tronarp20} and our experiments).

The integral that underlies $\epsilon_n$ can usually not be computed in closed form but needs to be approximated by a numerical integration scheme.
We use Bayesian quadrature (BQ) \citep{briol19}.
Not only does it fit neatly into the probabilistic framework, but it also allows us to place quadrature nodes freely in each domain $[t_n, t_{n+1})$.
If viewed as an integral from 0 to 1, we choose quadrature nodes at $0, 0.33, 0.5, 0.67, 1$.
These locations include the boundary points of the domains (0 and 1), as well as the nodes that will be introduced in case the error is too large (either 0.5, or 0.33 and 0.67).
This has the advantage that, at the start of the next iteration, we reuse the evaluation of the posterior at the new mesh points.
If the residual estimates the error, there is another advantage. Since the IEKS approximation is a minimum of a constrained optimisation task, the residual is zero at the boundaries of each interval. In this case, the integral is only computed on the three interior nodes.
For the same reasons, non-probabilistic solvers with residual control usually employ Gauss--Lobatto schemes \citep{kierzenka2001bvp}.

A final motivation for BQ is that we can tailor an integration kernel to $e$.
For instance the following reproducing kernel Hilbert spaces (RKHSs) are known \citep{tronarp20}: (i) the RKHS of $\nu$-times integrated Wiener process priors $Y(\cdot)$ is the Sobolev space of $(\nu+1)$-times weakly differentiable functions; (ii)
under some regularity assumptions on the ODE vector field, as well as on the (assumed to be) unique solution of the ODE, the RKHS of the residual $\ell(Y)(\cdot)$ is the Sobolev space of $\nu$-times weakly differentiable functions.
Therefore, we base the BQ scheme on a $(\nu-1/2)$th order Matérn prior, which has the same native space as the residual \citep{wendland04} (we use an exponentiated quadratic kernel for $\nu > 3$ because the required kernel embeddings are easier to compute \citep[Appendix J]{briol2015probabilistic}).

\begin{figure}
  \includegraphics{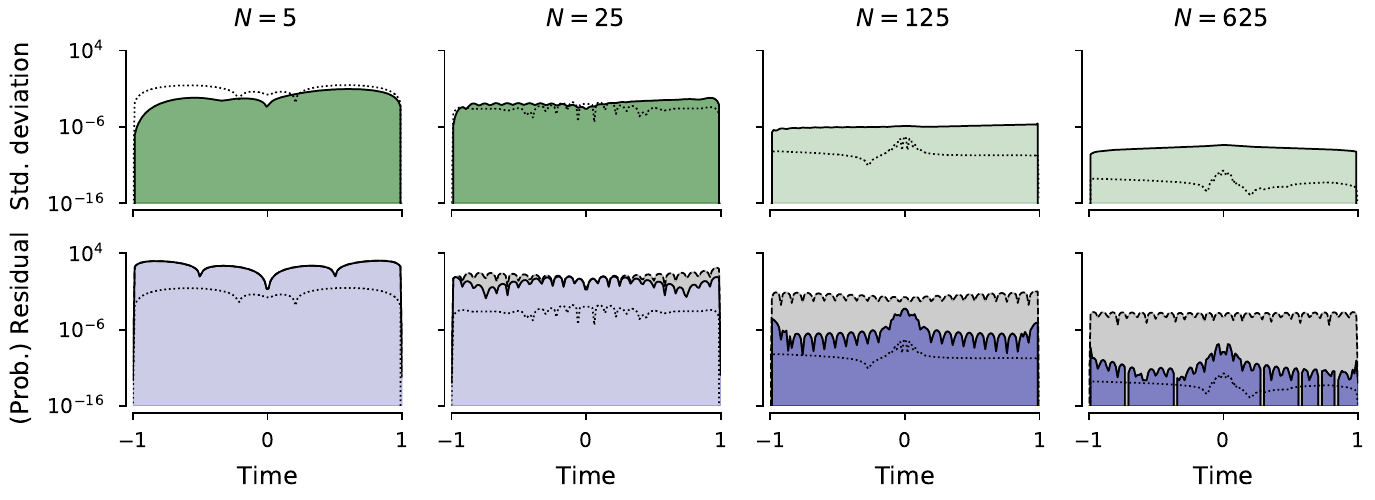}
  \caption{\textit{Error estimation on the seventh testproblem in \citep{mazzia2014test}.}
    Evaluated at $N=5$ (\figureposition{black}{left}), $N=25$ (\figureposition{black}{centre left}), $N=125$ (\figureposition{black}{centre right}), and $N=625$ equidistant grid points (\figureposition{black}{right}).
    Standard deviation (\figureposition{green!50!black}{top row}) and residual (\figureposition{blue!50!black}{bottom row}) respectively the probabilistic residual (\figureposition{gray}{bottom row}).
    True error in black. A good estimate accurately measures the magnitude of the error as well as the location of large deviation. On few points, the latter is less important so the well-calibrated standard deviation provides a good estimate. On many points, it is underconfident. Since for large $N$, the location of the error becomes increasingly important, the residual should be used; the probabilistic residual is consistently underconfident.
    The ``winners'' of each column have a darker colour.
  }
  \label{fig:error_estimates}
\end{figure}

Which one is the most reliable error estimate?
As a first testbed, we use the seventh in a collection of test problems for BVP solvers by \citet{mazzia2014test} (which will feature heavily in the remainder of this work).
The derivative of the solution of this linear BVP approaches a singularity if a specific parameter is chosen sufficiently small (we use $10^{-3}$). This poses challenges for error estimators and mesh-refinement strategies.
The error estimates are visualised in Figure \ref{fig:error_estimates}.
They suggest that at high tolerances, the standard deviation is more accurate than the residual; at low tolerances, the situation is reversed.
This trend is preserved when moving to more challenging setups (see Section \ref{sec:work_precision}).

To conclude, the probabilistic framework introduces three options for error estimation and comes with a natural algorithm to compute accumulated errors in BQ.
With everything explained so far, we can solve BVPs with an algorithm that adaptively refines the mesh when the solution is not sufficiently accurate.
After each mesh refinement, the iteration is restarted.
While it may be clear that the initial guess for the new IEKS implementation should be the approximate posterior from the previous computation, beginning a new Gauss--Newton scheme offers the chance to update the choice of other hyperparameters and thus set up a more appropriate probabilistic model for free (Section \ref{sec:calibration_em}).

\section{Calibration of hyperparameters with maximum-likelihood and EM}
\label{sec:calibration_em}

Thus far, an approximate BVP solution has been computed with $\sigma$, $m_0$, and $C_0$ set to default values.
Maximum-likelihood estimates of these hyperparameters can be computed by coordinate ascent, which repeats alternating updates
\begin{subequations}
  \begin{align}
    \sigma^\text{new}              & :=
    \argmax_\sigma
    \log p(\ell_L, \ell_{0:N}, \ell_R \,|\, \sigma, m_0^\text{new}, C_0^\text{new}),\label{eq:mle_diffusion}
    \\
    m_0^\text{new}, C_0^\text{new} & :=
    \argmax_{m_0, C_0}
    \log p(\ell_L, \ell_{0:N}, \ell_R \,|\, \sigma^\text{new}, m_0, C_0), \label{eq:mle_initial_parameters}
  \end{align}
\end{subequations}
until some stopping criterion is satisfied \citep{wright2015coordinate}.
A quasi-maximum likelihood update for $\sigma^\text{new}$ (Equation \ref{eq:mle_diffusion}) is available in closed form as a by-product of the forward-pass of each IEKS iteration.
This is also true for the specific order of updates detailed previously in Section \ref{sec:gaussian_bridge} (Proposition \ref{prop:quasi_mle} below).
\begin{proposition} \label{prop:quasi_mle}
  Assume that the initial covariance and the diffusion of the Wiener process depend multiplicatively on the scalar $\sigma^2$ (recall Equation \eqref{eq:gauss_markov_sde}).
  If $\ell_L$, $\ell_R$, and $\ell_{0:N}$ are noise-free (which herein they always are), the covariance of the posterior process depends multiplicatively on $\sigma^2$ and a quasi-maximum likelihood estimate for $\sigma$ is available in closed form.
\end{proposition}
The proof of this proposition is similar to the proof of Proposition 4 of \citet{tronarp19} yet requires a few additional manipulations because of the boundary value information contained in the bridge. A derivation -- and the precise formula for the quasi-MLE -- are in Appendix \ref{app:quasi_ml}.

\begin{figure}[h]
  \centering
  \includegraphics{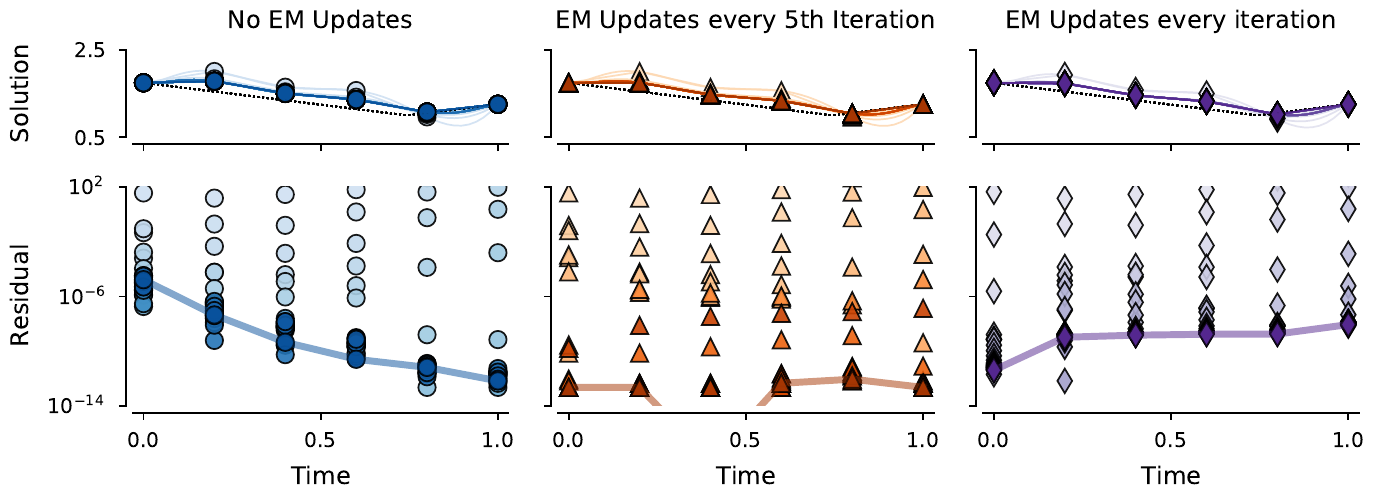}
  \caption{\textit{EM helps the IEKS overcome unknown initial conditions.}
    Depicted are a fixed total of the first 25 IEKS iterations (light to dark in each respective colour) on $N=6$ grid points, initialised with an EKS using a 7-times integrated Wiener process bridge prior on the 20th test problem in \citep{mazzia2014test}.
    Without any EM updates to the initial condition, the convergence of the IEKS is inhibited (\figureposition{blue!50!gray}{left}).
    EM updates every fifth IEKS iteration lead to the residual converging to zero reliably (\figureposition{orange!75!black}{centre}).
    Too many EM updates are not optimal either (\figureposition{violet!75!black}{right}).
  }
  \label{fig:em_improves_ieks}
\end{figure}

While $\sigma$ is tuned with quasi-maximum likelihood estimation, the parameters $m_0$ and $C_0$ of the initial distribution are separately calibrated with {a single step} of the expectation-maximisation (EM) algorithm \citep{dempster1977maximum, shumway1982approach} whenever the mesh needs to be refined, which implies a restart of the IEKS.
In other words, this ``outer loop'' around calls to the IEKS is \emph{already} part of the computational budget; therefore, sensible updates to the initial distribution parameters $m_0$ and $C_0$ are free.
The general idea of EM is to maximise a lower bound of \Cref{eq:mle_initial_parameters} instead of maximising it directly, by computing alternating $E$- and $M$-steps.
For parameter estimates in state-space models, the $E$-step of the EM algorithm is the posterior distribution in \Cref{eq:full_posterior} (see e.g. \citep{neal1998view}), a Gaussian approximation of which is available through the IEKS: recall $Y_\text{\tiny MAP}(t) \sim \Ncal( m_\text{\tiny MAP}(t),  C_\text{\tiny MAP}(t))$.
The $M$-step consists of \citep[Theorem 12.5 and Algorithm 12.7]{sarkka2013bayesian}
\begin{subequations}\label{eq:m_step}
  \begin{align}
    m_0^\text{new}          & =  m_\text{\tiny MAP}(t_0)                                                                                    \\
    \sigma^2 C_0^\text{new} & = \sigma^2 C_\text{\tiny MAP}(t_0) + (m_0^\text{new} - m_0^\text{old})(m_0^\text{new} - m_0^\text{old})^\top.
  \end{align}
\end{subequations}
EM steps always increase the likelihood, and for exponential families, convergence to a stationary point of the likelihood function is guaranteed \citep{wu1983convergence, shumway1982approach}.
Thus, computing alternating $E$- and $M$-steps until convergence (which we do not do) would eventually yield a good estimate of the parameters. But already in the pre-asymptotic regime and for a fixed total number of IEKS iterations, making an EM update every few steps helps convergence of the IEKS in subsequent iterations (Figure \ref{fig:em_improves_ieks}).

\section{The solver converges quickly on test problems}
\label{sec:work_precision}
\begin{figure}
  \centering
  \includegraphics{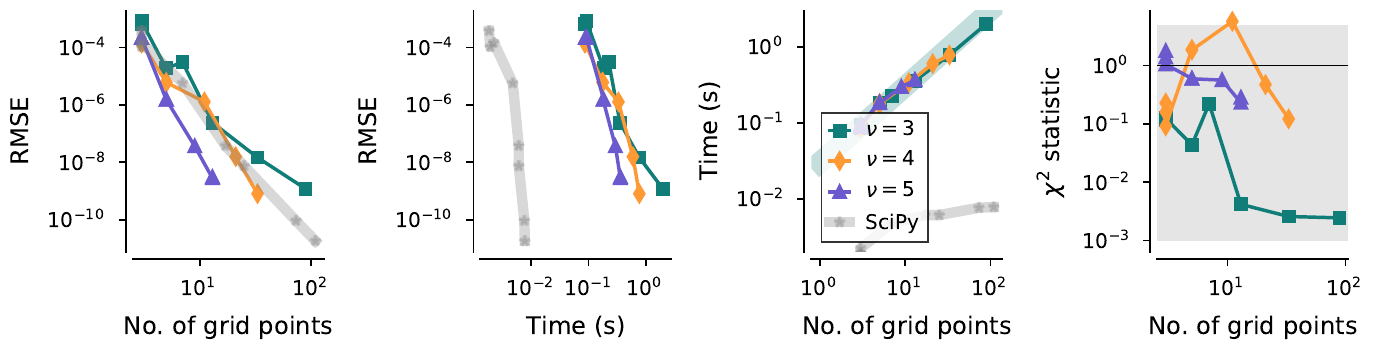}
  \caption{\textit{Results on Bratu's problem.}
    The higher-order solvers converge at least as fast as the SciPy reference (\figureposition{black}{left}), and are roughly by factor $\sim$ 100 slower (\figureposition{black}{centre left}, \figureposition{black}{centre right}; linear complexity reference line in the background). The $\chi^2$-statistic remains within $95\%$ confidence (\figureposition{black}{right}; intervals shaded in gray, mean ($=1$) in black). To show mesh refinement, the initial grid consisted of only three points; the probabilistic solver initialises with EKS and bridge, and uses the standard deviation as an error estimate.
  }
  \label{fig:bratu_workprecision}
\end{figure}

\begin{figure}
  \centering
  \includegraphics{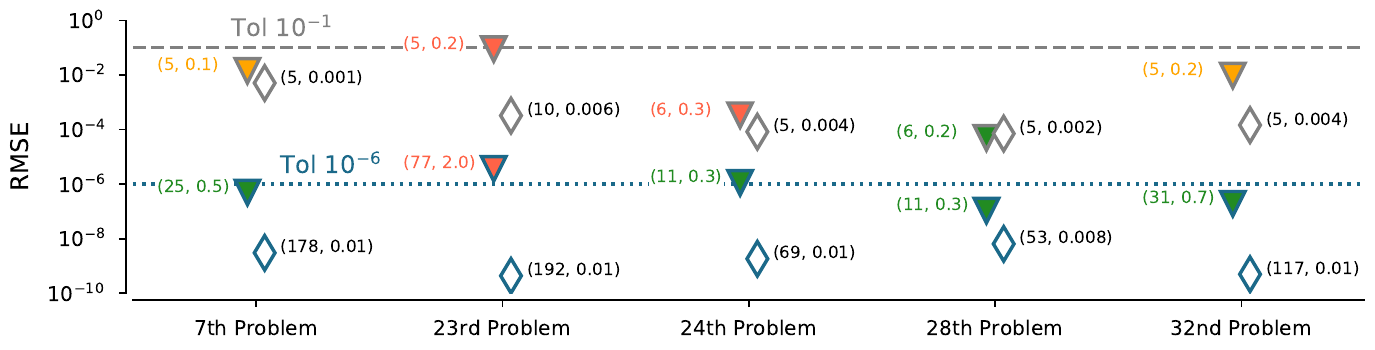}
  \caption{\textit{The solver efficiently computes (mostly) calibrated posteriors on many problems.}
    Probabilistic solver ($\triangledown$; $\nu = 6$) versus SciPy's BVP solver ($\lozenge$). Markers are annotated with the number of grid points and runtime (in seconds). The tolerances are $10^{-1}$ (\figureposition{gray}{gray}) and $10^{-6}$ (\figureposition{blue!50!gray}{blue}). The closer a coloured marker is to its reference line, the better. The fewer grid points and the less time required the better. Fill-color describes calibration: $\chi^2$ is within 80 $\%$ (\figureposition{green!50!black}{green}),  within $99 \%$ (\figureposition{orange}{orange}), or outside of these ranges (\figureposition{red!75!black}{red}). SciPy does not allow a notion of calibration.
  }
  \label{fig:problem7_workprecision}
\end{figure}

Now that all parts are in place, we evaluate the performance of the solver on a range of scenarios.
An efficient probabilistic numerical method should provide both a good point estimate (through its posterior mean) and error estimate (through its posterior covariance). First, the approximation error should decrease rapidly with the number of grid points; we report root-mean-square errors -- the lower, the better. Second, the width of the posterior distribution should be representative of the numerical approximation error (which has, to some extent, been shown in Section \ref{sec:mesh_refinement} already); we use the $\chi^2$-statistic \citep{bar2004estimation}.
If it is close to 1, the posterior uncertainty is calibrated.
A simulation of Bratu's problem \citep{bratu1914equations} for varying tolerances and orders $\nu$ suggests that the solver performs well in both metrics (Figure \ref{fig:bratu_workprecision}).
Reassuringly, higher orders of the solver lead to faster convergence, which motivates the analysis of convergence rates akin to the analysis of \citet{tronarp20} for initial value problem solvers.
The experiments also suggest that the uncertainties are calibrated but tend to be under-confident.
Efficient mesh refinement and fast convergence are evident when considering a wider range of test problems. Figure \ref{fig:problem7_workprecision} depicts the results of simulating five BVPs (all from \citet{mazzia2014test}): the 7th problem approaches a singularity in its derivative, the 23rd problem has a boundary layer at $t_\text{max}$, the 24th problem describes a fluid mechanical model of a shock wave, the 28th problem has a corner layer at $t_\text{min}$, and the 32nd problem involves fourth-order derivatives.
On all problems, the probabilistic solver efficiently computes calibrated posteriors at specified tolerances.
\section{Related work}
\label{sec:summary_markov_property_fast}
How does the proposed algorithm fit into the context of state-of-the-art probabilistic and non-probabilistic BVP solvers?
Headway on the probabilistic solution of BVPs has been made by \citet{hennig2014probabilistic}, \citet{arvanitidis2019fast}, and \citet{john19}.
\citet{hennig2014probabilistic} and \citet{arvanitidis2019fast} focus on the application of BVP solvers to Riemannian statistics.
None of the three algorithms exploit the state space structure of the prior with its beneficial computational complexity, nor are they concerned with error estimation, mesh refinement, and the other computational aspects to the extent that this work is.
In terms of accuracy and cost, the present approach should rather be compared to off-the-shelf non-probabilistic BVP solvers: for instance, those implemented in Matlab \citep{shampine1997matlab, kierzenka2001bvp, kierzenka2008bvp}, Python/SciPy \citep{virtanen2020scipy}, and Julia \citep{rackauckas2017differentialequations}.
These toolboxes contain algorithms that implement collocation formulas and gain linear-time complexity from sparse system matrices.
The Markov property makes our algorithm equally fast (in terms of the number of grid points $N$)  (Table \ref{tbl:computational_aspects}).
\begin{table}
  \centering
  \caption{Comparison of probabilistic and non-probabilistic BVP solvers.
  }
  \label{tbl:computational_aspects}
  \begin{tabular}{lll}
    \toprule
                               & Non-probabilistic         & Probabilistic (present work)          \\
    \midrule
    $O(N)$ achieved by         & Sparse matrices           & Markov property                       \\
    Error estimates            & Residual (point estimate) & Many options, e.g. standard deviation \\
    Initial guess              & Mandatory                 & Optional                              \\
    Uncertainty quantification & No                        & Yes                                   \\
    \bottomrule
  \end{tabular}
\end{table}
\begin{wrapfigure}{r}{0.45\textwidth}
  \begin{minipage}{0.45\textwidth    }
    \begin{algorithm}[H]
      \SetAlgoLined
      \KwIn{BVP, mesh, order ($\nu$), tolerances.}
      \KwOut{Probabilistic BVP Solution }
      Initialise with bridge and ODE filter\;
      \While{$\exists \geq 1$ interval with large error}{
        Run IEKS\;
        Update $m_0$ and $C_0$ (\Cref{eq:m_step})\;
        Update $\sigma$ \;
        Compute error between gridpoints\;
        Refine mesh where necessary\;
      }
      \caption{BVP Solver}
      \label{alg:bvp_solver}
    \end{algorithm}
  \end{minipage}
\end{wrapfigure}
The computational complexity of Algorithm \ref{alg:bvp_solver} is $O(I_\text{Mesh}I_\text{IEKS} N \nu^3d^3)$, where $I_\text{IEKS}$ is the number of IEKS iterations, and $I_\text{Mesh}$ is the number of mesh refinements.
In our experiments, we found $I_\text{IEKS}$ to be small, usually bounded by $10$.
The mesh refinement is designed to make $I_\text{Mesh}$ as small as possible.
Linear complexity in $N$ stems from the state space implementation of the IEKS and could potentially be reduced to $\log N$ by temporal parallelisation \citep{yaghoobi2021parallel}.
The cubic complexity in $\nu$ and in $d$ stems from the matrix-matrix operations that are required in a Kalman filter step  \citep{sarkka2013bayesian}.
Cubic complexity in $d$ suggests that high-order BVPs should be solved directly, without transforming them into first-order. This is not uncommon for BVP solvers \citep[Section 5.6]{ascher1995numerical} and is used herein (a version of $\ell_n$ that is suitable to high order ODEs is explained in Appendix \ref{app:solve_higher_order_odes_directly}).

\section{Conclusion}
\label{sec:conclusion}

We have arguably provided the first \emph{practically usable} probabilistic BVP solver. Our method achieves the same linear computational complexity as off-the-shelf solvers, with high-quality point estimates and calibrated uncertainty. Algorithmic parameters can be set automatically by the method, including some that have to be set manually for non-probabilistic solvers. Our method thus closes a methodological gap in the toolbox of probabilistic numerics.

\section*{Acknowledgements}

The authors gratefully acknowledge financial support by the German Federal Ministry of Education and Research (BMBF) through Project ADIMEM (FKZ 01IS18052B).
They also gratefully acknowledge financial support by the European Research Council through ERC StG Action 757275 / PANAMA; the DFG Cluster of Excellence ``Machine Learning - New Perspectives for Science'', EXC 2064/1, project number 390727645; the German Federal Ministry of Education and Research (BMBF) through the Tübingen AI Center (FKZ: 01IS18039A); and funds from the Ministry of Science, Research and Arts of the State of Baden-Württemberg.
Moreover, the authors thank the International Max Planck Research School for Intelligent Systems (IMPRS-IS) for supporting Nicholas Krämer.
The authors thank Nathanael Bosch, Filip Tronarp, and Jonathan Schmidt for valuable discussions. They are grateful to Georgios Arvanitidis for helping with the implementation of the manifold example in Figure \ref{fig:bvp_example}.

\bibliography{bibfile}

\begin{thebibliography}{44}
\providecommand{\natexlab}[1]{#1}
\providecommand{\url}[1]{\texttt{#1}}
\expandafter\ifx\csname urlstyle\endcsname\relax
  \providecommand{\doi}[1]{doi: #1}\else
  \providecommand{\doi}{doi: \begingroup \urlstyle{rm}\Url}\fi

\bibitem[Ascher et~al.(1995)Ascher, Mattheij, and Russell]{ascher1995numerical}
Uri~M Ascher, Robert~MM Mattheij, and Robert~D Russell.
\newblock \emph{Numerical Solution of Boundary Value Problems for Ordinary
  Differential Equations}.
\newblock SIAM, 1995.

\bibitem[Carmo(1992)]{carmo1992riemannian}
Manfredo Perdigao~do Carmo.
\newblock \emph{Riemannian Geometry}.
\newblock Birkh{\"a}user, 1992.

\bibitem[Arvanitidis et~al.(2018)Arvanitidis, Hansen, and
  Hauberg]{arvanitidis2017latent}
Georgios Arvanitidis, Lars~Kai Hansen, and S{\o}ren Hauberg.
\newblock Latent space oddity: On the curvature of deep generative models.
\newblock In \emph{International Conference on Learning Representations}, 2018.

\bibitem[Kersting et~al.(2020{\natexlab{a}})Kersting, Sullivan, and
  Hennig]{kersting19}
Hans Kersting, Tim~J Sullivan, and Philipp Hennig.
\newblock Convergence rates of {Gaussian} {ODE} filters.
\newblock \emph{Statistics and Computing}, 30\penalty0 (6):\penalty0
  1791--1816, 2020{\natexlab{a}}.

\bibitem[Tronarp et~al.(2021)Tronarp, S\"arkk\"a, and Hennig]{tronarp20}
Filip Tronarp, Simo S\"arkk\"a, and Philipp Hennig.
\newblock {B}ayesian {ODE} solvers: The maximum a posteriori estimate.
\newblock \emph{Statistics and Computing}, 31\penalty0 (23), 2021.

\bibitem[Bosch et~al.(2021)Bosch, Hennig, and Tronarp]{bosch2020calibrated}
Nathanael Bosch, Philipp Hennig, and Filip Tronarp.
\newblock Calibrated adaptive probabilistic {ODE} solvers.
\newblock In \emph{International Conference on Artificial Intelligence and
  Statistics}, 2021.

\bibitem[Kr{\"a}mer and Hennig(2020)]{kramer2020stable}
Nicholas Kr{\"a}mer and Philipp Hennig.
\newblock Stable implementation of probabilistic {ODE} solvers.
\newblock \emph{arXiv:2012.10106}, 2020.

\bibitem[Kersting et~al.(2020{\natexlab{b}})Kersting, Kr{\"a}mer, Schiegg,
  Daniel, Tiemann, and Hennig]{kersting2020differentiable}
Hans Kersting, Nicholas Kr{\"a}mer, Martin Schiegg, Christian Daniel, Michael
  Tiemann, and Philipp Hennig.
\newblock Differentiable likelihoods for fast inversion of 'likelihood-free'
  dynamical systems.
\newblock In \emph{International Conference on Machine Learning},
  2020{\natexlab{b}}.

\bibitem[Schmidt et~al.(2021)Schmidt, Kr{\"a}mer, and
  Hennig]{schmidt2021probabilistic}
Jonathan Schmidt, Nicholas Kr{\"a}mer, and Philipp Hennig.
\newblock A probabilistic state space model for joint inference from
  differential equations and data.
\newblock \emph{arXiv:2103.10153}, 2021.

\bibitem[Hennig and Hauberg(2014)]{hennig2014probabilistic}
Philipp Hennig and S{\o}ren Hauberg.
\newblock Probabilistic solutions to differential equations and their
  application to {Riemannian} statistics.
\newblock In \emph{International Conference on Artificial Intelligence and
  Statistics}, 2014.

\bibitem[Arvanitidis et~al.(2019)Arvanitidis, Hauberg, Hennig, and
  Schober]{arvanitidis2019fast}
Georgios Arvanitidis, Soren Hauberg, Philipp Hennig, and Michael Schober.
\newblock Fast and robust shortest paths on manifolds learned from data.
\newblock In \emph{International Conference on Artificial Intelligence and
  Statistics}, 2019.

\bibitem[John et~al.(2019)John, Heuveline, and Schober]{john19}
David John, Vincent Heuveline, and Michael Schober.
\newblock {GOODE}: A {G}aussian off-the-shelf ordinary differential equation
  solver.
\newblock In \emph{International Conference on Machine Learning}, 2019.

\bibitem[Tronarp et~al.(2019)Tronarp, Kersting, S\"arkk\"a, and
  Hennig]{tronarp19}
Filip Tronarp, Hans Kersting, Simo S\"arkk\"a, and Philipp Hennig.
\newblock Probabilistic solutions to ordinary differential equations as
  non-linear {B}ayesian filtering: A new perspective.
\newblock \emph{Statistics and Computing}, 29\penalty0 (6):\penalty0
  1297--1315, 2019.

\bibitem[Calderhead et~al.(2009)Calderhead, Girolami, and
  Lawrence]{calderhead2009accelerating}
Ben Calderhead, Mark Girolami, and Neil~D Lawrence.
\newblock Accelerating {B}ayesian inference over nonlinear differential
  equations with {G}aussian processes.
\newblock In \emph{Advances in Neural Information Processing Systems}, 2009.

\bibitem[Wenk et~al.(2020)Wenk, Abbati, Osborne, Sch{\"o}lkopf, Krause, and
  Bauer]{wenk2020odin}
Philippe Wenk, Gabriele Abbati, Michael~A Osborne, Bernhard Sch{\"o}lkopf,
  Andreas Krause, and Stefan Bauer.
\newblock {ODIN}: {ODE}-informed regression for parameter and state inference
  in time-continuous dynamical systems.
\newblock In \emph{AAAI Conference on Artificial Intelligence}, 2020.

\bibitem[Bell(1994)]{bell1994iterated}
Bradley~M Bell.
\newblock The iterated {K}alman smoother as a {Gauss--Newton} method.
\newblock \emph{SIAM Journal on Optimization}, 4\penalty0 (3):\penalty0
  626--636, 1994.

\bibitem[Knoth(1989)]{knoth1989globalization}
O~Knoth.
\newblock A globalization scheme for the generalized {Gauss--Newton} method.
\newblock \emph{Numerische Mathematik}, 56\penalty0 (6):\penalty0 591--607,
  1989.

\bibitem[Kierzenka and Shampine(2001)]{kierzenka2001bvp}
Jacek Kierzenka and Lawrence~F Shampine.
\newblock A {BVP} solver based on residual control and the {Matlab} {PSE}.
\newblock \emph{ACM Transactions on Mathematical Software}, 27\penalty0
  (3):\penalty0 299--316, 2001.

\bibitem[sci()]{scipybvpsolverdocs}
{SciPy's} \texttt{solve$\_$bvp} documentation.
\newblock
  \url{https://docs.scipy.org/doc/scipy/reference/generated/scipy.integrate.solve_bvp.html}.
\newblock Accessed: May 23, 2021.

\bibitem[mat()]{matlabbvpsolverdocs}
{Matlab's} \texttt{bvpinit} documentation.
\newblock \url{https://uk.mathworks.com/help/matlab/ref/bvpinit.html}.
\newblock Accessed: May 23, 2021.

\bibitem[jul()]{juliabvpsolverdocs}
{Julia's} \texttt{BoundaryValueProblem} documentation (in
  \texttt{DifferentialEquations.jl}).
\newblock \url{https://diffeq.sciml.ai/stable/types/bvp_types/}.
\newblock Accessed: May 23, 2021.

\bibitem[S{\"a}rkk{\"a}(2013)]{sarkka2013bayesian}
Simo S{\"a}rkk{\"a}.
\newblock \emph{Bayesian Filtering and Smoothing}.
\newblock Cambridge University Press, 2013.

\bibitem[S{\"a}rkk{\"a} and Solin(2019)]{sarkka2019applied}
Simo S{\"a}rkk{\"a} and Arno Solin.
\newblock \emph{Applied Stochastic Differential Equations}.
\newblock Cambridge University Press, 2019.

\bibitem[Mazzia(2014)]{mazzia2014test}
F~Mazzia.
\newblock Test set for boundary value problem solvers, release 0.5, 2014.

\bibitem[Kierzenka and Shampine(2008)]{kierzenka2008bvp}
Jacek Kierzenka and Lawrence~F Shampine.
\newblock A {BVP} solver that controls residual and error.
\newblock \emph{Journal of Numerical Analysis, Industrial and Applied
  Mathematics}, 3\penalty0 (1-2):\penalty0 27--41, 2008.

\bibitem[Briol et~al.(2019)Briol, Oates, Girolami, Osborne, and
  Sejdinovic]{briol19}
Fran{\c{c}}ois-Xavier Briol, Chris~J Oates, Mark Girolami, Michael~A Osborne,
  and Dino Sejdinovic.
\newblock Probabilistic integration: A role in statistical computation?
\newblock \emph{Statistical Science}, 34\penalty0 (1):\penalty0 1--22, 2019.

\bibitem[Wendland(2004)]{wendland04}
Holger Wendland.
\newblock \emph{Scattered Data Approximation}.
\newblock Cambridge University Press, 2004.

\bibitem[Briol et~al.(2015)Briol, Oates, Girolami, Osborne, and
  Sejdinovic]{briol2015probabilistic}
Fran{\c{c}}ois-Xavier Briol, Chris~J Oates, Mark Girolami, Michael~A Osborne,
  and Dino Sejdinovic.
\newblock Probabilistic integration.
\newblock \emph{arXiv:1512.00933v1}, 2015.

\bibitem[Wright(2015)]{wright2015coordinate}
Stephen~J Wright.
\newblock Coordinate descent algorithms.
\newblock \emph{Mathematical Programming}, 151\penalty0 (1):\penalty0 3--34,
  2015.

\bibitem[Dempster et~al.(1977)Dempster, Laird, and Rubin]{dempster1977maximum}
Arthur~P Dempster, Nan~M Laird, and Donald~B Rubin.
\newblock Maximum likelihood from incomplete data via the {EM} algorithm.
\newblock \emph{Journal of the Royal Statistical Society: Series B
  (Methodological)}, 39\penalty0 (1):\penalty0 1--22, 1977.

\bibitem[Shumway and Stoffer(1982)]{shumway1982approach}
Robert~H Shumway and David~S Stoffer.
\newblock An approach to time series smoothing and forecasting using the {EM}
  algorithm.
\newblock \emph{Journal of Time Series Analysis}, 3\penalty0 (4):\penalty0
  253--264, 1982.

\bibitem[Neal and Hinton(1998)]{neal1998view}
Radford~M Neal and Geoffrey~E Hinton.
\newblock A view of the {EM} algorithm that justifies incremental, sparse, and
  other variants.
\newblock In \emph{Learning in Graphical Models}, pages 355--368. Springer,
  1998.

\bibitem[Wu(1983)]{wu1983convergence}
CF~Jeff Wu.
\newblock On the convergence properties of the {EM} algorithm.
\newblock \emph{The Annals of Statistics}, pages 95--103, 1983.

\bibitem[Bar-Shalom et~al.(2004)Bar-Shalom, Li, and
  Kirubarajan]{bar2004estimation}
Yaakov Bar-Shalom, X~Rong Li, and Thiagalingam Kirubarajan.
\newblock \emph{Estimation With Applications to Tracking and Navigation: Theory
  Algorithms and Software}.
\newblock John Wiley \& Sons, 2004.

\bibitem[Bratu(1914)]{bratu1914equations}
Gh~Bratu.
\newblock Sur les {\'e}quations int{\'e}grales non lin{\'e}aires.
\newblock \emph{Bulletin de la Soci{\'e}t{\'e} Math{\'e}matique de France},
  42:\penalty0 113--142, 1914.

\bibitem[Shampine and Reichelt(1997)]{shampine1997matlab}
Lawrence~F Shampine and Mark~W Reichelt.
\newblock The {Matlab ODE} suite.
\newblock \emph{SIAM Journal on Scientific Computing}, 18\penalty0
  (1):\penalty0 1--22, 1997.

\bibitem[Virtanen et~al.(2020)Virtanen, Gommers, Oliphant, Haberland, Reddy,
  Cournapeau, Burovski, Peterson, Weckesser, Bright, et~al.]{virtanen2020scipy}
Pauli Virtanen, Ralf Gommers, Travis~E Oliphant, Matt Haberland, Tyler Reddy,
  David Cournapeau, Evgeni Burovski, Pearu Peterson, Warren Weckesser, Jonathan
  Bright, et~al.
\newblock {SciPy} 1.0: Fundamental algorithms for scientific computing in
  {Python}.
\newblock \emph{Nature Methods}, 17\penalty0 (3):\penalty0 261--272, 2020.

\bibitem[Rackauckas and Nie(2017)]{rackauckas2017differentialequations}
Christopher Rackauckas and Qing Nie.
\newblock {DifferentialEquations.jl}--a performant and feature-rich ecosystem
  for solving differential equations in {Julia}.
\newblock \emph{Journal of Open Research Software}, 5\penalty0 (1), 2017.

\bibitem[Yaghoobi et~al.(2021)Yaghoobi, Corenflos, Hassan, and
  S{\"a}rkk{\"a}]{yaghoobi2021parallel}
Fatemeh Yaghoobi, Adrien Corenflos, Sakira Hassan, and Simo S{\"a}rkk{\"a}.
\newblock Parallel iterated extended and sigma-point {Kalman} smoothers.
\newblock In \emph{IEEE International Conference on Acoustics, Speech and
  Signal Processing}, 2021.

\bibitem[Nocedal and Wright(2006)]{nocedal2006numerical}
Jorge Nocedal and Stephen Wright.
\newblock \emph{Numerical Optimization}.
\newblock Springer, 2006.

\bibitem[Immer(2020)]{immer2020disentangling}
Alexander Immer.
\newblock Disentangling the {Gauss--Newton} method and approximate inference
  for neural networks.
\newblock \emph{arXiv:2007.11994}, 2020.

\bibitem[Higham(2002)]{higham2002accuracy}
Nicholas~J Higham.
\newblock \emph{Accuracy and Stability of Numerical Algorithms}.
\newblock SIAM, 2002.

\bibitem[Rasmussen and Williams(2006)]{rasmussen2003gaussian}
Carl~E Rasmussen and Christopher~KI Williams.
\newblock \emph{Gaussian Processes for Machine Learning}.
\newblock MIT Press, 2006.

\bibitem[Schweppe(1965)]{schweppe1965evaluation}
Fred Schweppe.
\newblock Evaluation of likelihood functions for {G}aussian signals.
\newblock \emph{IEEE transactions on Information Theory}, 11\penalty0
  (1):\penalty0 61--70, 1965.

\end{thebibliography}

\appendix
\numberwithin{equation}{section}
\section{IEKS as a Gauss--Newton Laplace approximation}
\label{app:ieks_as_laplace}
The present appendix shows that the iterated extended Kalman filter (IEKS) yields a Gauss--Newton Laplace approximation of the posterior distribution (recall Equation \eqref{eq:map_problem_optimisation} from the main paper)
\begin{align}\label{app:posterior_distribution_ell}
  p(Y(\Tbb) \mid \ell_L(Y) = 0, ~ \ell_R(Y) = 0, ~ \ell_{0:N}(Y) = 0).
\end{align}
Let $\delta$ be the Dirac delta.
Introduce the random variables $Z_L$, $Z_R$, $Z_{0:N}$ as
\begin{align}
  Z_L \mid Y(t_0) \sim \delta(\ell_L(Y)), \quad Z_R \mid Y(t_\text{max}) \sim \delta(\ell_R(Y)), \quad Z_n \mid Y(t_n) \sim \delta(\ell_n(Y)(t_n)).
\end{align}
The posterior distribution in Equation \eqref{app:posterior_distribution_ell} becomes
\begin{align}\label{eq:posterior_distribution_Z}
  p(Y(\Tbb) \mid Z_L = 0, ~ Z_R = 0,~ Z_{0:N} = 0).
\end{align}
The difference between Equation \eqref{app:posterior_distribution_ell} and Equation \eqref{eq:posterior_distribution_Z} is only notational.
The reformulation in terms of $Z$ will be useful in the next step.
The prior distribution is Gaussian (recall Section \ref{sec:generative_model}),
\begin{align}
  p(Y(\Tbb)) = \Ncal(m(\Tbb), K(\Tbb, \Tbb))
\end{align}
for mean and covariance functions $m$ and $K$ that correspond directly to the stochastic differential equation (SDE) representation in Equation \ref{eq:gauss_markov_sde} in the main paper \citep[Chapter 12]{sarkka2019applied}.
The use of this general Gaussian process formulation, as opposed to a sequential notion that exploits the Markov property, will simplify the notation (and simultaneously slightly generalise the result).

To show that the IEKS provides a Gauss--Newton version of the Laplace approximation, it is instructive to consider a relaxed version of the Dirac likelihoods; that is, let $\lambda > 0$ and (re)define (recall the ODE $\dot y = f(y, t)$, and the boundary conditions $Ly(t_0)=y_0$, $R y(t_\text{max})=y_\text{max}$),
\begin{subequations}
  \begin{align}
    Z_L \mid Y(t_0)          & \sim \Ncal(L Y_0(t_0) - y_0), \lambda I),                  \\
    Z_R \mid Y(t_\text{max}) & \sim \Ncal(R Y_0(t_\text{max}) - y_\text{max}, \lambda I), \\
    Z_n \mid Y(t_n)          & \sim \Ncal(Y_1(t_n) - f(Y_0(t_n), t_n), \lambda I).
  \end{align}
\end{subequations}
In these formulas, $I$ is always an identity matrix of appropriate size.
The limit $\lambda \rightarrow 0$ recovers the Dirac likelihoods used in the previous paragraph and in the paper.
In the Gaussian relaxation, the MAP estimate is the argument that minimises the following objective (the negative log-posterior),
\begin{align}
  \arg \min_{Y(\Tbb)} ~ \frac{1}{2} \Vcal_1(Y(\Tbb)) + \frac{1}{2}\Vcal_2(Y(\Tbb))+ \frac{1}{2}\Vcal_3(Y(\Tbb)),
\end{align}
which uses the abbreviations
\begin{subequations}
  \begin{align}
    \Vcal_1(Y(\Tbb)) & := \|m(\Tbb) - Y(\Tbb)\|_{K(\Tbb, \Tbb)^{-1}}^2,                                                           \\
    \Vcal_2(Y(\Tbb)) & := \frac{1}{\lambda} \|L Y_0(t_0) - y_0\|^2 + \frac{1}{\lambda}\| R Y_0(t_\text{max}) - y_\text{max})\|^2, \\
    \Vcal_3(Y(\Tbb)) & :=  \frac{1}{\lambda} \sum_{n=0}^N \| Y_1(t_n) - f(Y_0(t_n), t_n)\|^2 .
  \end{align}
\end{subequations}
$\Vcal_1$ captures the prior distribution, $\Vcal_2$ the boundary conditions, and $\Vcal_3$ the (artificial) ODE ``measurements''.
Only $\Vcal_3$ includes non-linearities.

Let $\xi=(\xi_0, ..., \xi_N) \in \Rbb^{(N+1) \times d(\nu + 1)}$ (there are $N+1$ points in $\Tbb$) be the result of a previous Gauss--Newton iteration (or the initialisation, respectively).
Denote by $P_0$ and $P_1$ the projection matrices from $Y$ to $Y_0$, and $Y$ to $Y_1$, respectively.
Gauss--Newton optimisers such as the IEKS iteratively linearise the non-linearities of the objective ``inside the norm'' at $\xi$, and solve the resulting linear least-squares problem in closed form \citep{nocedal2006numerical}.
In other words, let
\begin{align}
  f(y, t_n) \approx f(P_0\xi_n, t_n) + \nabla f(P_0\xi_n, t_n)(y - P_0\xi_n)
\end{align}
be the first-order Taylor series linearisation of $f$ at $P_0\xi_n$ (each grid-point  $t_n$ uses a different $\xi_n$).
Then, the IEKS minimises
\begin{align} \label{eq:batch_objective}
  \Vcal(Y(\Tbb)) := \frac{1}{2} \|m(\Tbb) - Y(\Tbb)\|_{K(\Tbb, \Tbb)^{-1}}^2 + \frac{1}{2\lambda}\|F Y(\Tbb) + b\|^2,
\end{align}
which uses the batch notation (abbreviate $F_n := P_1 - \nabla f(P_0\xi_n, t_n)P_0$),
\begin{align}
  F = \begin{pmatrix}
    L      & 0      & \hdots & 0      \\
    0      & \hdots & \hdots & R      \\
    F_0    & 0      & \hdots & 0      \\
    0      & F_1    & \ddots & \vdots \\
    \vdots & \ddots & \ddots & \vdots \\
    0      & \hdots & \hdots & F_N
  \end{pmatrix}, \quad
  b = \begin{pmatrix}
    y_0                                        \\
    y_\text{max}                               \\
    \nabla f(\xi_0, t_0) \xi_0 - f(\xi_0, t_0) \\
    \nabla f(\xi_1, t_1) \xi_1 - f(\xi_1, t_1) \\
    \vdots                                     \\
    \nabla f(\xi_N, t_N) \xi_N - f(\xi_N, t_N)
  \end{pmatrix}.
\end{align}
This is a linear least-squares problem and can be solved in closed form with GP regression -- or, as in the present setting, with a Kalman smoother \citep{bell1994iterated}.
The mean of this solution becomes the new iterate $\xi$.
Unless a fixed point has been found, the procedure is repeated.

The Hessian of the objective in Equation \eqref{eq:batch_objective}\footnote{We call the Hessian of the Gauss--Newton objective the Gauss--Newton Hessian of the full objective \citep{immer2020disentangling}.} and its inverse are
\begin{subequations}
  \begin{align}
    \nabla^2 \Vcal(x)        & := K(\Tbb, \Tbb)^{-1} + \frac{1}{\lambda} F^\top F,
    \\
    (\nabla^2 \Vcal(x))^{-1} & := K(\Tbb, \Tbb) - K(\Tbb, \Tbb) F^\top (F K(\Tbb, \Tbb) F^\top + \lambda)^{-1} F K(\Tbb, \Tbb).
  \end{align}
\end{subequations}
The functional form of the inverse is revealed by, for instance, the matrix inversion lemma \citep{higham2002accuracy}.
The inverse Hessian identifies as the posterior covariance of Gaussian process regression (respectively the Kalman smoother) \citep{rasmussen2003gaussian,sarkka2019applied}.

At the final iteration of the IEKS, the objective is linearised at the MAP estimate ($\xi = m_\text{\tiny \scshape MAP}$).
This fixed point then yields a Gaussian approximation of the posterior, where the mean is the MAP estimate, and the covariance is the negative inverse Gauss--Newton Hessian of the log-posterior, evaluated at the MAP estimate.
This shows how the IEKS yields a Gauss--Newton Laplace transform of the relaxed objective.
The limit $\lambda \rightarrow 0$ translates this to the Dirac objectives used in the paper.

In summary, the IEKS yields a Gauss--Newton Laplace approximation of the posterior, because (i) it uses a Gauss--Newton approximation of the non-linear objective, which (ii) can then be solved in closed form with a Kalman smoother, which -- since it delivers Gaussian posteriors -- is (iii) its own Laplace approximation.
Put differently, each iteration of the IEKS yields a Gaussian approximation of the posterior, where the covariance is the negative inverse Hessian of the log-posterior, evaluated at the mean -- when the mean converges to the MAP estimate, this makes the IEKS compute a Laplace approximation.

\section{Transition densities of the bridge prior are available in closed form}
\label{app:transition_densities_bridge}

The present section describes the transition densities of the bridge prior.
Recall the SDE representation of the prior process $Y$ (Equation \eqref{eq:gauss_markov_sde} in the main paper).
Due to the Markov property, the law of $Y$ factorises as
\begin{align}
  p(Y(\Tbb)) = p(Y(t_0)) \prod_{n=1}^N p(Y(t_n) \mid Y(t_{n-1})).
\end{align}
The initial distribution
\begin{align}
  p(Y(t_0)) = \Ncal(m_0, \sigma^2 C_0)
\end{align}
is part of the prior model (Section \ref{sec:generative_model} in the main paper).
The transition densities
\begin{align}
  p(Y(t_{n+1}) \mid Y(t_n)) = \Ncal(\Phi(t_{n+1}, t_n) Y(t_n), \sigma^2Q(t_{n+1}, t_n))
\end{align}
use the definitions \citep{sarkka2019applied},
\begin{subequations}
  \begin{align}
    \Phi(t, s) & := \exp(A(t-s))                                                                                               \\
    Q(t, s)    & := \int_0^{t-s} \Phi(t, \tau) B B^\top \Phi(s, \tau)^\top \diff \tau . \label{eq:prior_process_noise_formula}
  \end{align}
\end{subequations}
Both quantities can be computed with, e.g., matrix fractions \citep{sarkka2019applied}.
$A$ and $B$ stem from the SDE representation (Equation \eqref{eq:gauss_markov_sde} in the main paper).
The process noise covariance is of the form $\sigma^2 Q(t, s)$ because the diffusion of the Wiener process is (by assumption) $\Gamma = \sigma^2 I$ (and the diffusion of the Wiener process would enter Equation \eqref{eq:prior_process_noise_formula} as $B B^\top \rightsquigarrow B \Gamma B^\top$ \citep{sarkka2019applied}).

\subsection{Initial distribution}
The first objective of the present section is the parametrisation of the updated initial distribution (recall the shorthand for the information sources, first introduced in Equation \eqref{eq:map_estimate_with_shorthand} in the main paper)
\begin{align}
  p(Y(t_0) \mid \ell_L, \ell_R).
\end{align}
It arises as follows.
The joint distribution is
\begin{align}
  p(Y(t_0), \ell_L, \ell_R) = \Ncal(\xi, \sigma^2 \Xi),
\end{align}
with
\begin{align}
  \xi & := \begin{pmatrix}
    m_0         \\
    L m_0 - y_0 \\
    R \Phi(t_\text{max}, t_0) m_0 - y_\text{max}
  \end{pmatrix}
  , \quad
  \Xi := \begin{pmatrix}
    \Xi_1      & \Xi_2      & \Xi_3 \\
    \Xi_2^\top & \Xi_4      & \Xi_5 \\
    \Xi_3^\top & \Xi_5^\top & \Xi_6
  \end{pmatrix},
\end{align}
where we abbreviated
\begin{subequations}
  \begin{align}
    \Xi_1 & := C_0,                                                                                                  \\
    \Xi_2 & := C_0 L^\top,                                                                                           \\
    \Xi_3 & := C_0 \Phi(t_\text{max}, t_0)^\top R^\top,                                                              \\
    \Xi_4 & :=  L C_0 L^\top,                                                                                        \\
    \Xi_5 & :=L C_0 \Phi(t_\text{max}, t_0)^\top R^\top,                                                             \\
    \Xi_6 & := R \left[\Phi(t_\text{max}, t_0)C_0 \Phi(t_\text{max}, t_0)^\top + Q (t_\text{max}, t_0)\right]R^\top.
  \end{align}
\end{subequations}
Mean and covariance of $Y(t_0)$ conditioned on $\ell_L$ and $\ell_R$ now follow from standard conditioning rules of Gaussian distributions \citep{rasmussen2003gaussian}.
Since $C_0$ and the process noise of the covariance depend multiplicatively on $\sigma$, so does $\Xi$.

\subsection{Transition densities}
Let $Y(t_n) \sim \Ncal(m_n, \sigma^2 C_n)$ and recall $\ell_L$ and $\ell_R$.
The second objective of the present section is the transition density from $Y(t_{n-1})$ to $Y(t_n)$ under acknowledgement of the boundary conditions.
The joint distribution of $Y(t_{n+1})$ and the right-hand side boundary condition, given $Y(t_n)$, is
\begin{align}
  p(Y(t_{n+1}), \delta_R \mid Y(t_n)) = \Ncal( \zeta, \sigma^2 \Lambda)
\end{align}
with
\begin{align}
  \zeta =
  \begin{pmatrix}
    \Phi(t_{n+1}, t_n) m_n \\
    R \Phi(t_\text{max}, t_n) m_n - y_\text{max}
  \end{pmatrix}
  , \quad \text{and} \quad
  \Lambda =
  \begin{pmatrix}
    \Lambda_{1} & \Lambda_{2}^\top \\
    \Lambda_{2} & \Lambda_3
  \end{pmatrix},\end{align}
which uses the abbreviations
\begin{subequations}
  \begin{align}
    \Lambda_1 & :=
    \Phi(t_{n+1}, t_n) C_n\Phi(t_{n+1}, t_n)^\top + \sigma^2 Q(t_{n+1}, t_n) \\
    \Lambda_2 & :=
    \Lambda_1 \Phi(t_\text{max}, t_{n+1})^\top R^\top                        \\
    \Lambda_3 & :=
    R \left[ \Phi(t_\text{max}, t_{n+1}) \Lambda_1 \Phi(t_\text{max}, t_{n+1})^\top + \sigma^2 Q(t_\text{max}, t_{n+1}) \right] R^\top.
  \end{align}
\end{subequations}
Notably, since the covariance of $Y(t_n)$ depends multiplicatively on $\sigma^2$, all entries of $\Lambda$ do as well (this will be useful in Appendix \ref{app:quasi_ml}).
Finally, the distribution $p(Y(t_{n+1}) \mid \ell_R, Y(t_n))$ is Gaussian with mean and covariance that are available with the usual conditioning formula for multivariate Gaussians \citep{rasmussen2003gaussian}.
On a side note: $\Lambda_3$ is ill-conditioned for $t_\text{max} \approx t_{n+1}$, which is a problem that can be solved with appropriate preconditioning as well as square-root implementation \citep{kramer2020stable}.

\section{The quasi-MLE is essentially unaffected by the bridge }
\label{app:quasi_ml}
The present section proves that the quasi-maximum likelihood estimate (quasi-MLE) for the diffusion $\sigma$ is available in closed form, even for the bridge prior.
A formula is given as well.
We say that a matrix $X$ depends multiplicatively on $\sigma^2$, if it satisfies $X = \sigma^2 \breve X$ for some $\breve X$.
First, we need to establish that all the covariances that contribute to the (approximate) prediction error decomposition depend multiplicatively on $\sigma^2$. This has partly been done in Appendix \ref{app:transition_densities_bridge}.
Second, this multiplicative dependency gives rise to a closed-form solution for the quasi-MLE.

\citet{tronarp20} establish that for a conventional Gauss--Markov prior, and noise-free ODE measurements, the covariances of the predictive distribution $p(Y(t_{n+1} \mid Y(t_n))$ depend multiplicatively on $\sigma^2$.
Appendix \ref{app:transition_densities_bridge} established the same for the predictive distribution $p(Y(t_{n+1}) \mid \delta_R, Y(t_n))$ of the bridge.

The same will hold not only for the predictive distribution, but also for the filtering covariances, as shown next.
The (iterated) extended Kalman filter approximates the non-linear ODE likelihood \citep{tronarp19,sarkka2013bayesian}
\begin{align}
  p(\ell_n \mid Y(t_n)) = \delta ( Y_1(t_n) - f(Y_0(t_n), t_n))
\end{align}
with a first order Taylor approximation around some $\xi_n \in \Rbb^{d(\nu + 1)}$ (recall from Appendix \ref{app:transition_densities_bridge} that $P_0$ is the projection matrix from $Y$ to $Y_0$),
\begin{align}
  p(\ell_n \mid Y(t_n)) \approx \delta(Y_1(t_n) - \nabla f(P_0 \xi_n, t_n) (Y_0(t_n) - P_0\xi_n)).
\end{align}
For the non-iterated Kalman filter, $\xi_n$ is the mean of the predictive distribution \citep{sarkka2013bayesian}; for the iterated extended Kalman smoother, $\xi_n$ is the mean of the previous iteration \citep{bell1994iterated}.
Since this is a noise-free (i.e. Dirac) likelihood, the law of $\ell_n$ given $Y(t_n)$ is Gaussian with a covariance that depends multiplicatively on $\sigma^2$.
Therefore, the covariance of $Y(t_{n+1})$ conditioned on $\ell_n = 0$ (approximately, with an [iterated] extended Kalman filter), depends multiplicatively on $\sigma^2$ as well.

Consider the following take on the prediction error decomposition \citep{schweppe1965evaluation},\footnote{It is not the traditional prediction error decomposition in so far as it employs the bridge prior.}
\begin{align}\label{eq:new_prediction_error_decomposition}
  p(\ell_L, \ell_{0:N}, \ell_R \mid \sigma) =
  p(\ell_0\mid\ell_L, \ell_R, \sigma)
  p(\ell_R \mid \ell_L, \sigma )
  p(\ell_L \mid \sigma)
  \prod_{n=1}^N p(\ell_n \mid \ell_{n-1}, \ell_R, \sigma)
\end{align}
which mirrors the factorisation of the prior in \Cref{eq:prior_bridge_distribution} of the main paper.
All of the terms in \Cref{eq:new_prediction_error_decomposition} are (approximated by) Gaussian distributions (which has been shown above),
\begin{subequations}
  \begin{align}
    p(\ell_L \mid \sigma)                     & = \Ncal(z_L, \sigma^2 S_L)         \\
    p(\ell_R \mid \ell_L, \sigma )            & = \Ncal(z_R, \sigma^2 S_R)         \\
    p(\ell_0\mid\ell_L, \ell_R, \sigma)       & \approx   \Ncal(z_0,\sigma^2  S_0) \\
    p(\ell_n \mid \ell_{n-1}, \ell_R, \sigma) & \approx \Ncal(z_n, \sigma^2 S_n)
  \end{align}
\end{subequations}
either because they are Gaussian by construction (the boundary conditions are linear), or because the (iterated) extended Kalman filter employs a Gaussian approximation.
The joint likelihood of $\ell_L$, $\ell_R$, and $\ell_{0:N}$ is maximised by the term that minimises the negative log-probability of all of these random variables being zero (neglecting some additive constants that do not depend on $\sigma$),
\begin{align}\label{eq:likelihood_quasi_mle}
  - 2\log p(\ell_L, \ell_{0:N}, \ell_R \mid \sigma) & \approx
  \frac{1}{\sigma^2} \Psi_0 + \log (\sigma^2 ) \Psi_1   + \text{const}
\end{align}
which employs the abbreviations
\begin{align}
  \Psi_0   :=
  z_L^\top  S_L^{-1} z_L  +z_R^\top  S_R^{-1} z_R +  \sum_{n=0}^N  z_n^\top  S_n^{-1} z_n, \quad
  \Psi_1   :=
  d_L + d_R  + d(N+1).
\end{align}
Setting the derivative of the likelihood in Equation \eqref{eq:likelihood_quasi_mle} with respect to $\sigma$ to zero, yields
\begin{align}
  -\frac{2}{\sigma^3} \Psi_0 + \frac{2}{\sigma} \Psi_1 = 0
  \quad \Leftrightarrow \quad
  {\sigma^2}= \frac{\Psi_0}{\Psi_1}
\end{align}
which gives a formula for the quasi-maximum likelihood estimate of the diffusion.
From the first iteration of the IEKS onwards, this quasi-MLE equals the quasi-MLE from \citet{tronarp19}; for the initialisation via the extended Kalman smoother, the bridge prior alters the linearisation over the law of $\ell_n$, and thus affects the quasi-MLE.

\section{Solve higher-order BVPs directly}
\label{app:solve_higher_order_odes_directly}
The present appendix explains how BVPs based on higher-order ODEs can be solved directly without transforming them into first-order problems.
Many problems in the test set by \citet{mazzia2014test} are second-order.
The 32nd problem in \citep{mazzia2014test} (which features in Section \ref{sec:work_precision} of the main paper) is fourth-order.

As an instructive example, consider the second order ODE
\begin{align}
  \ddot y(t) = f(\dot y(t), y(t), t).
\end{align}
If $y$ and $\dot y$ would be stacked into a new state $z := (\dot y, y)$, the ODE could equivalently be written as
\begin{align}
  \dot z(t) = g(z(t), t),
\end{align}
with $g(z(t)) = f(\dot y(t), y(t))$.
Recall from Equation \eqref{eq:likelihood_ode} in the main paper that such first-order ODEs give rise to the information operator
\begin{align}
  \ell_{1\text{st}}(Y)(t) :=  Y_1(t) - f(Y_0(t), t).
\end{align}
With this approach, higher-order ODEs can be solved.
The increased dimensionality of the ODE problem makes this inefficient (as outlined in Section \ref{sec:conclusion}).

ODE information operators can straightforwardly be generalised to second-order ODEs, via
\begin{align}
  \ell_{2\text{nd}}(Y)(t) :=  Y_2 - f(Y_1(t), Y_0(t), t).
\end{align}
Higher-order ODEs, like the fourth-order ODE that has been part of the experiments, use the same concept:
provided $\nu \geq 4$, we can define a likelihood for fourth-order ODEs,
\begin{align}
  \ell_{4\text{th}}(Y)(t) :=  Y_4 - f(Y_3(t), Y_2(t),Y_1(t), Y_0(t), t).
\end{align}
The only requirement for this to work is that $\nu$ is sufficiently large.
All of these likelihood functions can be used inside an extended Kalman filter.
Solving higher-order BVPs directly, without transforming them into first-order problems, is not uncommon for BVP solvers \citep[Section 5.6]{ascher1995numerical}.



\end{document}